\documentclass[runningheads]{llncs}

\usepackage[T1]{fontenc}
\usepackage{amsmath,amssymb,amsfonts}
\usepackage{graphicx}
\graphicspath{{figures/}}
\usepackage{textcomp}
\usepackage{xcolor}
\usepackage{bm}
\usepackage{booktabs}
\usepackage{placeins}
\usepackage{url}
\def\BibTeX{{\rm B\kern-.05em{\sc i\kern-.025em b}\kern-.08em
    T\kern-.1667em\lower.7ex\hbox{E}\kern-.125emX}}
\begin{document}

\title{Subjective-Graph LLM Agents for Simulating Uncertainty in Classroom Social Perception}

\titlerunning{Subjective-Graph LLM Agents}

\author{Jinming Yang\inst{1}$^{*}$ \and
Xinyu Jiang\inst{2}$^{*}$ \and
Xinshan Jiao\inst{1}$^{\dagger}$ \and
Xinping Zhang\inst{3}}
\authorrunning{J. Yang et al.}
\institute{
Complex Lab, School of Computer Science and Engineering,
University of Electronic Science and Technology of China, Chengdu, China\\
\email{\{yangjinming,jiaoxinshan\}@std.uestc.edu.cn}
\and
Department of Computer Science, City University of Hong Kong, Hong Kong\\
\email{xjiang346-c@my.cityu.edu.hk}
\and
School of Health and Medical Technology, Chengdu Neusoft University, Chengdu, China\\
\email{ZhangXinping@nsu.edu.cn}\\[0.5ex]
$^{*}$Equal contribution. $^{\dagger}$Corresponding author.
}

\maketitle

\begin{abstract}
Social actors do not observe a common social world: each individual forms judgments from a partial and potentially distorted view of the surrounding network. We study whether graph-local evidence and credibility-weighted communication can generate persistent distortions in perceived academic standing, even when agents repeatedly receive objective performance signals. We introduce a data-constrained multi-agent framework in which LLM agents operate through individualized subjective graphs that determine peer visibility, evidence access, and interaction opportunities. Agents exchange uncertainty-annotated assessments, evaluate message credibility, and maintain explicit Gaussian belief states updated through Bayesian fusion. We evaluate the framework on 12 middle-school classrooms comprising 482 students, using questionnaire-derived social information and six consecutive examinations. On the Social-Observed subset ($n=419$), collective ranking error increases from $0.066 \pm 0.008$ to $0.124 \pm 0.009$ across six epochs despite repeated exam-based anchoring. Ablations associate individualized visibility and LLM-based trust gating with more stable long-horizon behavior, while constrained retrieval primarily safeguards against global-information leakage. Compared with evaluated DeGroot configurations, the proposed framework achieves lower final ranking error; those DeGroot configurations exhibit near-zero terminal opinion diversity. These findings establish subjective-graph LLM agents as a mechanism-oriented framework for data-constrained simulated social perception. Code is available at \url{https://anonymous.4open.science/r/Rashomonomon-0126}.

\end{abstract}

\section{Introduction}
In educational psychology and computational social science, academic achievement is commonly understood as an outcome jointly shaped by individual ability, effort, and contextual factors~\cite{edm_success_2023,edm_2023_mq,simulation_based_academic_prediction_2024}. Accumulating empirical evidence, however, suggests that students' learning behaviors and academic performance do not develop in isolation; rather, they are embedded in classroom social networks that evolve over time~\cite{peer_networks_review_2024,systematic_review_edm_2023}. Peer interactions, information exchange, and social comparison can influence how individuals assess their own competence and others' levels, and these shifting perceptions may in turn affect learning-related decisions. As such processes unfold, performance differentiation and structural disparities can gradually emerge at the group level. Explaining how macro-level learning outcomes arise from micro-level social interaction remains a long-standing concern at the intersection of educational psychology and social computing.

From a modeling perspective, this problem is multi-actor, network-structured, and temporally evolving. Static prediction or single-time-point descriptions cannot capture how student judgments change within observed social structure~\cite{marl_survey_2023,agent_based_modelling_llm_2025}. Existing approaches largely fall into rule-based multi-agent models and data-driven predictors. Rule-based models are interpretable but often rely on simplified cognition and weak empirical constraints~\cite{rule_based_vs_llm_2024,ijcai_mas_survey_2024,beyond_static_responses_2025}. Data-driven predictors learn static mappings and do not naturally support sustained multi-step interaction with evolving cognition~\cite{gnn_social_rec_2023,inflect_dgnn_2023}. Large language models can generate complex social behavior and explanations~\cite{stanford_simulating_2023,llm_generative_abm_2024}, but multi-agent deployments still need mechanisms that enforce observed social structure without introducing global information~\cite{persona_aligned_2025}.

Building on these considerations, this paper introduces a large-language-model-driven probabilistic multi-agent framework for data-constrained classroom simulation. Under incomplete and unevenly distributed social information, different individuals can form distinct interpretations of the same group structure; localized interaction may then accumulate these differences into group-level misperception.

Unlike graph models with a global view, we construct an individualized subjective graph for each student agent. The graph constrains retrieval, peer visibility, interaction partners, and belief updates. Agents exchange uncertainty-annotated assessments, evaluate credibility with the language model, and update Gaussian belief states, enabling a mechanism-oriented analysis of simulated classroom social perception over time.

This paper makes the following key contributions:
\begin{itemize}
    \item \textbf{Subjective social modeling under partial observability.}
    We formulate social perception as a decentralized epistemic process and introduce individualized subjective graphs that jointly govern peer visibility, evidence accessibility, and interaction opportunities. This formulation replaces the conventional global-view assumption with agent-specific representations of the same social environment.

    \item \textbf{Explicit and uncertainty-aware belief propagation.}
    We develop a probabilistic agent architecture in which every agent--peer belief is represented by an explicit Gaussian state and updated through uncertainty-annotated communication, LLM-based credibility gating, and Bayesian fusion. The resulting framework makes both belief evolution and uncertainty propagation quantitatively traceable, rather than embedding them implicitly in natural-language memory.

    \item \textbf{Data-constrained evidence of collective misperception dynamics.}
    Using questionnaire-derived social information and six consecutive real-world examinations, we conduct a longitudinal simulation that connects local interaction mechanisms to emergent group-level outcomes. Comparisons with DeGroot dynamics, static predictors, and graph-based baselines distinguish locally constrained interaction from one-shot prediction and from classical consensus dynamics that exhibit near-zero terminal diversity.
\end{itemize}

Accordingly, we investigate the following research questions:

\textbf{RQ1:} Can graph-local, uncertainty-aware interaction generate persistent collective ranking distortion despite repeated performance anchors?

\textbf{RQ2:} How are individualized visibility, constrained retrieval, and LLM-based trust gating associated with the resulting long-horizon dynamics?

\textbf{RQ3:} How does the proposed mechanism differ from static predictors and classical consensus dynamics in final ranking behavior?
\section{Related Work}

\subsection{Multi-Agent Social Simulation and Opinion Dynamics}
Multi-agent social simulation has long served as a foundation for studying collective behavior and cognitive evolution. Early work was largely built on rule-driven agent models or classical opinion-dynamics frameworks, where information diffusion over networks is described through pre-specified interaction rules~\cite{degroot1974_reaching,friedkin1990_social}. Such formulations are analytically interpretable, yet they often rely on stylized cognitive assumptions and consequently struggle to capture heterogeneous perception in real social environments or to represent uncertainty as something agents can explicitly express~\cite{lorenz2007_continuous,golub2010_naive,abm_meets_genai_2024}. Subsequent studies introduced probabilistic representations or stochastic processes to account for fluctuations in judgment and the impact of noise. Even so, many of these models operate on abstract network structures and provide limited constraints from concrete social context~\cite{acemoglu2011_opinion,modeling_interpersonal_2024}.

\subsection{Predictive Graph Learning and LLM-Based Agents}
As deep learning matured, data-driven approaches were increasingly adopted to model social behavior and group-level outcomes---for instance, using regression or sequence models to predict individual actions or aggregate indicators~\cite{kipf2018_neural,battaglia2018_relational,gnn_influencer_2024}. While these methods can deliver strong predictive accuracy, they typically emphasize static input--output mappings and offer only limited support for explaining the mechanisms of cognitive evolution under sustained, multi-step interaction~\cite{gnn_social_rec_2023,inflect_dgnn_2023}. More recently, the natural language understanding and generation capabilities of large language models have accelerated their adoption in multi-agent systems. A growing body of work employs language models as agents' decision or generation modules to simulate phenomena such as dialogue, cooperation, and social bias~\cite{park2023_generative,stanford_simulating_2023,llm_generative_abm_2024}. These results suggest that language models can produce socially plausible behavior patterns with internal consistency and contextual dependence. At the same time, many existing designs assume that agents operate in a global---or near-global---information environment, leaving perceptual heterogeneity induced by social structure comparatively underexplored~\cite{persona_aligned_2025}.

\subsection{Belief Updating, Uncertainty, and Affective Factors}
In parallel, belief propagation and group cognition research has long examined how individual judgments accumulate through interaction and shape collective cognitive structure. Classical formulations typically update beliefs via linear or weighted aggregation rules, and some extensions introduce probabilistic beliefs to represent uncertainty~\cite{bayesian_teaching_llm_2026}. Within language-model-based multi-agent frameworks, however, belief updating is often embedded implicitly in generation behavior or implemented through heuristics, which makes it difficult to systematically characterize uncertainty as an explicit state and to study how its effects compound over time~\cite{huang2024_bayesagent,llm_bayes_update_2025,uncertainty_propagation_2025}. Psychological and affective factors further complicate this picture, as they are widely regarded as influential drivers of information processing and social judgment~\cite{gross1998_emotion,picard1997_affective}. Empirical findings indicate that emotional state, anxiety level, and self-perception can substantially alter how individuals receive, interpret, and evaluate information~\cite{pnas_cognitive_bias_2024,mitigating_bias_multiagent_2024}. Several multi-agent models have begun to incorporate emotion- or psychology-related variables as agent attributes to simulate information distortion and judgment volatility observed in real social settings. In most cases, however, these factors are treated as static perturbations, and the coupling between affective states, perceived social structure, and belief diffusion remains insufficiently explored~\cite{modeling_interpersonal_2024,llm_agent_mental_health_2025}.

\section{Method}
\label{sec:method}

\begin{figure}[t]
    \centering
    \includegraphics[width=\textwidth]{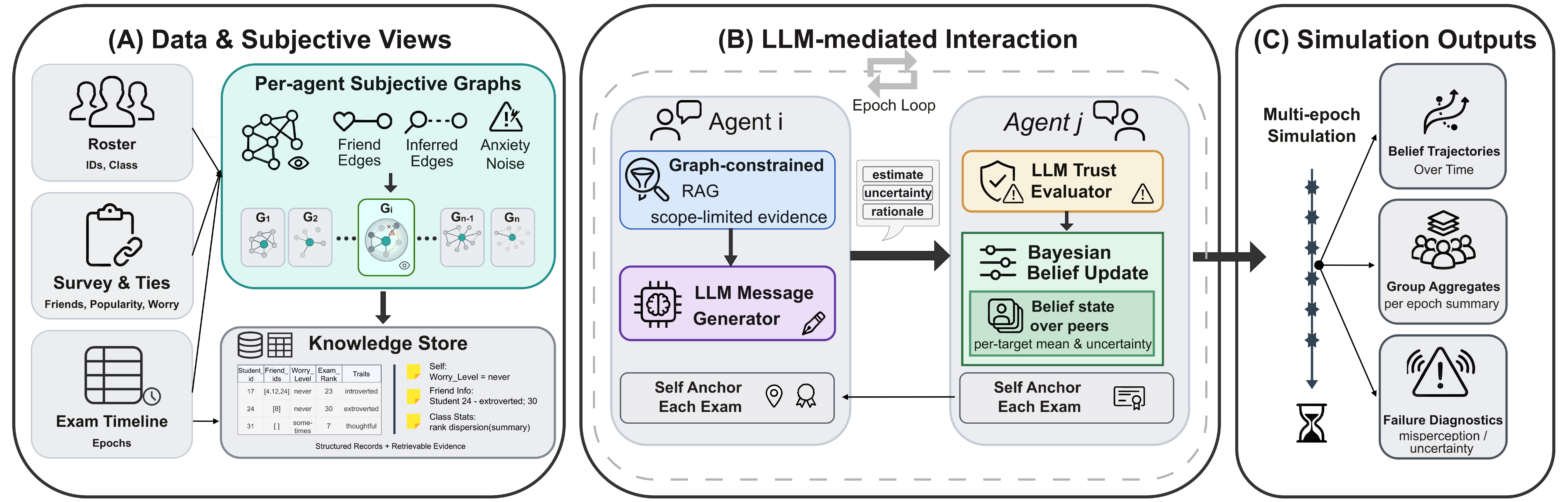}
    \caption{Overview of the proposed subjective-graph LLM-agent framework: (A) individualized subjective graphs and retrievable local knowledge; (B) graph-constrained RAG messaging with LLM trust gating and Bayesian updates; (C) multi-epoch outputs and diagnostics.}

    \label{fig:overall_framework}
\end{figure}

Consider a classroom system composed of $N$ students, with node set $V=\{1,\ldots,N\}$. The system evolves on a discrete timeline $t\in\{1,\ldots,6\}$, where each time step corresponds to one real examination. Let $Y_i^{(t)}$ denote the observed exam score of student $i$ at time $t$, which induces the ground-truth ranking $R^{(t)}=\mathrm{Rank}(Y^{(t)})$. We also introduce a one-dimensional latent variable $\theta_i^{(t)}$ to represent the student's underlying academic ability (or level of performance) on a unified scale and write $\Theta^{(t)}=\{\theta_i^{(t)}\}_{i\in V}$.

We model each student as an agent driven by a large language model. Operating under a restricted informational view, an agent generates judgments and interaction behaviors intended to approximate cognition and decision-making in a real classroom social environment. For each peer $k$, student agent $j$ maintains a subjective belief distribution $B_{jk}^{(t)}(\theta)$ over ability at time $t$; this distribution encodes both a point estimate and its associated uncertainty.

Social interaction is abstracted as noisy message exchange. During the round of $r$-th interaction at time $t$, a sender $i$ transmits a message $m_{i\rightarrow j}^{(t,r)}$ to a receiver $j$. The message provides assessments of the academic ability or social standing of one or more target students, together with an explicit expression of uncertainty. The receiver then updates its beliefs as a function of the credibility of the message~\cite{uncertainty_propagation_2025}. Given the observed score sequence $\{Y^{(t)}\}_{t=1}^6$ and limited questionnaire-derived information, we study how a system governed by local visibility, noisy interactions, trust gating, and belief fusion generates individual belief states over time, and how these states in turn induce group-level ranking errors and internal uncertainty diagnostics. To structure the exposition, we organize the method into three coupled mechanisms: subjective structure and visibility constraints, interaction with trust gating, and belief updating with the definition of macro-level quantities, as illustrated in Fig.~\ref{fig:overall_framework}.


\subsection{Subjective Graph}

Students do not share a unified understanding of the classroom social structure. To capture such perceptual heterogeneity, we define for each student agent $i$ an individualized subjective social graph $G_i=(V,E_i,\pi_i)$. Here, $E_i\subseteq V\times V$ denotes the set of social ties that student $i$ believes to exist or to be reachable, and $\pi_i(e)\in[0,1]$ assigns a subjective reachability (or credibility) weight to an edge $e$~\cite{friedkin1990_social}. Importantly, $G_i$ is not intended to reconstruct the true classroom network; instead, it specifies the social structure that the agent can perceive and utilize when forming judgments and engaging in interactions~\cite{cognitive_social_structures_2023}. The construction of the subjective graph is fully constrained by observable social-relational and psychometric information from the questionnaire. Explicitly reported friendships define the agent's base visible neighborhood, while subjective judgments about peer relations and social standing provide a prior modulation of edge reachability weights.

To represent individual differences in the stability of social perception, we further introduce a noise parameter $\alpha_i$ mapped from questionnaire scales. In the implementation, explicitly reported friend edges are retained with probability $1-\alpha_i$, and each non-neighbor is added as a spurious edge with probability $0.1\alpha_i$. A larger $\alpha_i$ indicates higher uncertainty when assessing whether a social tie exists; consequently, even within the same classroom environment, different individuals may form subjective graphs with substantially different structures.

Beyond the subjective graph itself, we impose an explicit constraint on information access. We treat all classroom-related structured information as an external knowledge base $D$. For a judgment or interaction task performed by agent $i$ at time $t$, the accessible evidence is determined by a subjective-graph-constrained retrieval operator:
\begin{equation}
K_i^{(t)}=\mathrm{RAG}\!\left(i, q_i^{(t)}, D;\,G_i\right),
\end{equation}
where $q_i^{(t)}$ is the task-induced query~\cite{auto_rag_2024}. The retrieval process is only allowed to return local information consistent with the reachability defined by $G_i$, thereby ruling out implicit injection of global knowledge at the mechanism level~\cite{rag_survey_gao_2024}. Constrained retrieval is introduced primarily as a mechanism-level safeguard against global-information leakage, rather than as an accuracy-improving component. The retrieved evidence may contain omissions, false positives, or phrasing biases, with the perturbation strength modulated by $\alpha_i$; this design integrates individual differences in both \emph{visibility scope} and \emph{perceptual precision} within a single modeling component.

\subsection{Interaction and Trust}

Under constrained information access, social interaction is modeled as message exchange augmented with explicit uncertainty and credibility markers. Within each time step $t$, the system runs $R$ rounds of interaction to emulate information sharing during the interval between consecutive exams. In round $r$, agent $i$ samples up to $K$ interaction partners from its subjective neighborhood and sends a structured message $m_{i\rightarrow j}^{(t,r)}$ to receiver $j$. The message includes an evaluation of a target student $k$, denoted by $\hat{s}_{i\rightarrow k}^{(t,r)}$, together with self-reported uncertainty $\hat{u}_{i\rightarrow k}^{(t,r)}$, thereby separating \emph{what is claimed} from \emph{how confident the sender is in the claim}~\cite{uncertainty_propagation_2025}. A receiver does not necessarily accept a message at face value. We therefore introduce a trust-gating weight $\omega_{i\rightarrow j}^{(t,r)}\in[0,1]$ to represent receiver $j$'s subjective assessment of sender $i$'s credibility in the current context~\cite{trust_modeling_mas_2024}. The scalar $\omega_{i\rightarrow j}^{(t,r)}$ is produced by a large language model conditioned on constrained evidence and an agent-specific attribute summary. Since LLM-based evaluators may themselves exhibit systematic preference biases, this trust score is treated as a noisy gate rather than a ground-truth judgment~\cite{self_preference_bias_2026}. Given $\omega_{i\rightarrow j}^{(t,r)}$ and $\hat{u}_{i\rightarrow k}^{(t,r)}$, we instantiate the observation precision as $\tau_{i\rightarrow k}^{(t,r)}=\max\{0.01,\omega_{i\rightarrow j}^{(t,r)}(1-\hat{u}_{i\rightarrow k}^{(t,r)})/0.3^2\}$, so messages that are both more credible and less hesitant receive higher influence~\cite{bayesian_teaching_llm_2026}.

\subsection{Belief Update and Metrics}

To keep inference tractable, for any receiver $j$ and target $k$ we approximate the belief with a Gaussian distribution:
\begin{equation}
B_{jk}^{(t)}(\theta)=\mathcal{N}\!\left(\mu_{jk}^{(t)},\,v_{jk}^{(t)}\right),
\end{equation}
where $\mu_{jk}^{(t)}$ denotes the mean belief and $v_{jk}^{(t)}$ denotes the belief variance.
After $j$ receives a message from $i$ about $k$, the belief is updated via precision-weighted Bayesian fusion~\cite{llm_bayes_update_2025}:
\begin{equation}
\mu_{jk}' = \frac{(v_{jk})^{-1}\mu_{jk} + \tau_{i\to k}\hat{s}_{i\to k}}{(v_{jk})^{-1}+\tau_{i\to k}},\qquad
v_{jk}' = \frac{1}{(v_{jk})^{-1}+\tau_{i\to k}}.
\end{equation}

Each time step also provides an exogenous anchor through the observed exam outcomes. At the beginning of time $t$, the simulator calibrates agent $j$'s self-belief using its current self-performance anchor:
\begin{equation}
\mu_{jj}^{(t)} \leftarrow h(Y_j^{(t)}),\qquad v_{jj}^{(t)} \leftarrow v_{\text{self}}\ll 1
\end{equation}
where $h(\cdot)$ maps the current exam outcome to the unified ability scale, and $v_{\mathrm{self}}=0.05^2$ reflects high confidence in self-observed performance. In our implementation, $h$ uses the within-class rank-normalized score, $h(Y_j^{(t)})=1-(\mathrm{rank}_j^{(t)}-1)/(n_c-1)$, only as a simulator-side scaling operation; it is not interpreted as a claim that students observe the complete classroom ranking, and it does not expose other students' scores or full ordering through retrieval.

Aggregating individual beliefs induces group-level macro observables. Let
$\boldsymbol{\mu}_j^{(t)}=(\mu_{j1}^{(t)},\ldots,\mu_{jN}^{(t)})$,
and define the aggregated mean belief as
\begin{equation}
\bar{\mu}^{(t)} = \frac{1}{N}\sum_{j=1}^N \mu_{j\cdot}^{(t)} \in \mathbb{R}^N,
\end{equation}
This yields a group-perceived ranking
$\hat{R}^{(t)}=\mathrm{Rank}(\bar{\boldsymbol{\mu}}^{(t)})$.
We define the ranking error as
\begin{equation}
\mathrm{DPAE}(t)=1-\rho\!\left(\hat{R}^{(t)},R^{(t)}\right),
\end{equation}
where $\rho(\cdot,\cdot)$ is the Spearman rank correlation.
Let $T_k^{(t)}$ and $\hat{T}_k^{(t)}$ denote the top-$k$ student sets induced by the ground-truth ranking $R^{(t)}$ and the group-perceived ranking $\hat{R}^{(t)}$, respectively. We define Top-$k$ high-achiever identification accuracy as the overlap ratio:
\begin{equation}
\mathrm{Acc@}k(t)=\frac{\left|\hat{T}_k^{(t)}\cap T_k^{(t)}\right|}{k}.
\end{equation}
We maintain the mean belief variance
\begin{equation}
\mathrm{Unc}(t)=\frac{1}{N^2}\sum_{j=1}^{N}\sum_{k=1}^{N}v_{jk}^{(t)}
\end{equation}
as an internal uncertainty state of the simulation, but do not treat it as an independently validated empirical outcome.

\section{Experiments and Results}
This section evaluates the proposed subjective-graph LLM-agent framework on data-constrained middle-school classroom simulations. The experiments are designed to address a central question: when agents are restricted to local visibility and engage in noisy social interaction, what group-level misperception patterns can emerge in the simulation? Section~4.2 addresses RQ1, Section~4.3 addresses RQ2, Section~4.4 addresses RQ3, and Section~4.5 reports exploratory classroom heterogeneity.

\subsection{Experimental Settings}
\paragraph{Data}
We study $12$ middle-school classrooms in which all students have complete score records across $6$ consecutive examinations, yielding a total of $482$ students. Based on the availability of friend information in the questionnaire, we construct two subsets: (i) \textbf{Full-Temporal} ($n=482$), used to define the ground-truth score trajectories and ranking benchmarks; and (ii) \textbf{Social-Observed} ($n=419$), consisting of students who reported at least one friend ID and are therefore eligible for subjective-graph construction and interaction. Students not included in \textbf{Social-Observed} are treated as weakly observed boundary nodes: their beliefs are updated primarily via the exogenous self-score anchor, while they can still appear as passive targets of others' beliefs.

\paragraph{Protocol}
The system evolves over $6$ discrete time steps (epochs), each corresponding to a real examination. At the beginning of each time step, every agent first receives the self-performance calibration described above, after which the system executes $R=3$ rounds of social interaction. In each round, an agent samples up to $K=3$ interaction partners from the visible neighborhood of its subjective social graph. All stochasticity is controlled by fixed random seeds.

\paragraph{Implementation}
LLM calls use Llama-3.1-8B-Instruct with temperature $0.7$. Evaluation outputs are JSON triples of target, ability score, and uncertainty, while trust gating outputs JSON fields \texttt{trust\_weight} and \texttt{uncertainty}. RAG returns at most $8$ graph-reachable evidence items from self, friend, classmate, and class-statistic scopes. The \textbf{No-RAG} ablation keeps subjective-graph partner sampling but replaces peer-evidence retrieval with minimal self and class-statistic context. The self-anchor uses the rank-normalized score defined in Section~3.3 as an environment-side calibration, and $\alpha_i=\min(1,\mathrm{worry}_i/3)$ under the default noise scale. At epoch $t$, agents and retrieval access no future exam records or peer score table. Feature and graph baselines use five-fold cross-validation over questionnaire attributes and one-hop survey-friend structure.

\paragraph{Metrics}
We report DPAE and Acc@$k$ as evaluation metrics. For each seed, metrics are first computed per classroom and then averaged uniformly across the 12 classrooms; reported $\pm$ values are standard deviations across seeds, and figure bands are exploratory bootstrap intervals over seeds. Belief variance is maintained as an internal uncertainty state of the simulation but is not treated as an independently validated empirical outcome.

\subsection{RQ1: Temporal Evolution of Collective Misperception}
Baseline results are averaged over five seeds; all other ablation settings are run under the shared seed set $\{42,43,44\}$ to enable paired comparisons with the Baseline. Under the Baseline configuration, we examine how the deviation between the group-perceived ranking and the ground-truth ranking evolves over time across five seeds (seeds$=\{42,43,44,45,46\}$). The results show a directionally consistent accumulation of ranking misperception: $\mathrm{DPAE}$ increases from $0.066\pm0.008$ at the first exam to $0.124\pm0.009$ at the sixth exam, while the Spearman correlation decreases from $0.934\pm0.008$ to $0.876\pm0.009$. This pattern indicates that, even with an exogenous anchor provided by exam signals at every time step, noise and uncertainty propagated through localized interactions can accumulate into a structured, time-amplifying deviation at the group level, rather than manifesting as transient random fluctuations.

\begin{figure}[t]
\centering
\includegraphics[width=0.78\textwidth,trim=0 0 413 40,clip]{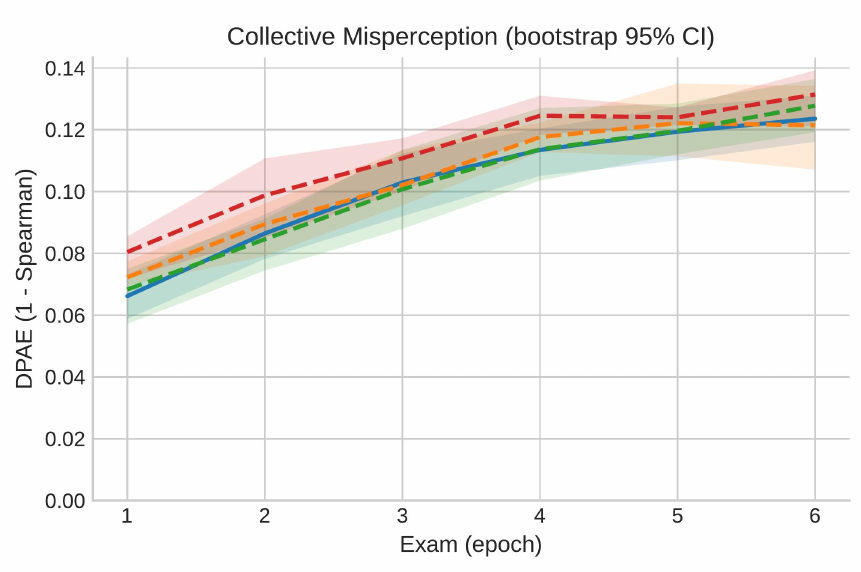}
\caption{Temporal $\mathrm{DPAE}$ trends for Baseline and ablations. Lower is better. Lines show across-seed means: solid blue is Baseline, dashed orange is No-RAG, dashed green is No-Subjective-Graph, and dashed red is No-LLM-Trust. Shaded bands show exploratory bootstrap intervals over the evaluated seeds (Baseline: five seeds; ablations: three seeds).}

\label{fig:temporal_dpae_spearman}
\end{figure}

As shown in Fig.~\ref{fig:temporal_dpae_spearman}, $\mathrm{DPAE}$ increases monotonically with exam index and the Spearman correlation declines in tandem, with consistent directionality across random seeds. The figure also overlays the ablation settings to show that the upward error trend is not unique to one configuration. Exploratory uncertainty intervals computed over the seed dimension shift upward over time, suggesting that the observed ``accumulated misperception'' is not driven by a single outlier run.

\subsection{RQ2: Ablation Study}

\begin{table}[t]
\centering
\caption{Baseline vs.\ ablations (mean$\pm$std). Subscripts 1/6 denote Epoch~1/6; Calls are total LLM calls (k).}
\label{tab:multiseed_ablation}
\footnotesize
\setlength{\tabcolsep}{3pt}
\resizebox{\textwidth}{!}{%
\begin{tabular}{lcccccc}
\toprule
Setting & seeds & DPAE$_1$ & DPAE$_6$ & $\rho_6$ & Acc@3$_6$ & Calls(k)\\
\midrule
Baseline & 5 & $\bm{0.066\pm0.008}$ & $0.124\pm0.009$ & $0.876\pm0.009$ & $\bm{0.278}$ & 535.8\\
No-RAG & 3 & $0.072\pm0.004$ & $\bm{0.121\pm0.011}$ & $\bm{0.879\pm0.011}$ & 0.269 & 294.2\\
No-Subjective-Graph & 3 & $0.068\pm0.008$ & $0.128\pm0.007$ & $0.872\pm0.007$ & 0.222 & 331.6\\
No-LLM-Trust & 3 & $0.080\pm0.006$ & $0.131\pm0.008$ & $0.869\pm0.008$ & $\bm{0.278}$ & 22.4\\
\bottomrule
\end{tabular}
}
\end{table}

To assess component behavior, we evaluate three ablations: (1) \textbf{No-RAG}, which disables graph-constrained peer-evidence retrieval while retaining minimal self and class-statistic context; (2) \textbf{No-Subjective-Graph}, which removes individual-specific visibility by forcing all agents to share a single visibility graph; and (3) \textbf{No-LLM-Trust}, which replaces the LLM-derived trust weight with a simple consistency function. The \textbf{No-LLM-Trust} variant is included to test whether trust gating is directionally associated with lower noise propagation over multiple epochs.

Table~\ref{tab:multiseed_ablation} reports performance at Epoch~1 and Epoch~6 (mean$\pm$standard deviation over random seeds). Removing individual visibility \textbf{No-Subjective-Graph} or trust gating \textbf{No-LLM-Trust} worsens final performance at Epoch~6, increasing $\mathrm{DPAE}$ and reducing correlation. Notably, \textbf{No-Subjective-Graph} lowers Acc@3 from $0.278$ to $0.222$, indicating that ignoring visibility heterogeneity weakens the group's ability to identify top-performing students in this simulation. By contrast, \textbf{No-RAG} achieves comparable or slightly better Epoch~6 ranking scores while substantially reducing the total number of calls. This result supports our interpretation that constrained retrieval mainly acts as a safeguard against global-information leakage, rather than as a direct accuracy-improving component. Across matched seeds, the paired effects for \textbf{No-Subjective-Graph} and \textbf{No-LLM-Trust} align in sign, but the three-seed ablations should be treated as exploratory.

\FloatBarrier
\subsection{RQ3: External Baselines and DeGroot Dynamics}
On the \textbf{Social-Observed} subset ($n=419$ across 12 classrooms), we compare a set of external baselines that do not rely on LLM-mediated interaction, and summarize their ranking performance at the final exam (Epoch~6) in Table~\ref{tab:external_baselines}. Overall, \textbf{Random} and \textbf{Self-Only} exhibit little to no meaningful correlation with the ground-truth ranking. Feature-based predictors---including linear regression and an MLP regressor trained on questionnaire-derived attributes---reach Spearman correlations of only about $0.16$, and their Top-3 identification rates remain low (Table~\ref{tab:external_baselines}). Static graph learning baselines (SGC/GCN-like and GAT-like, both using 1-hop aggregation) perform on a similar scale to these regressors, suggesting that a one-shot use of structure or static features is insufficient to account for group-level ranking judgments that evolve over time.

\begin{figure}[t]
\centering
\includegraphics[width=\textwidth]{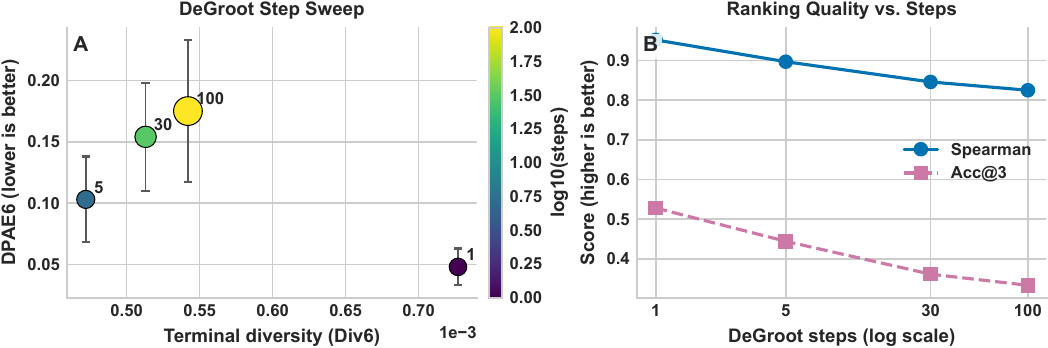}
\caption{DeGroot consensus dynamics under increasing update steps at Epoch~6. Panel A plots terminal diversity against DPAE, where lower DPAE is better and near-zero Div$_6$ indicates consensus collapse within DeGroot dynamics; labels denote DeGroot steps. Panel B shows that Spearman $\rho$ and Acc@3 decrease as the number of DeGroot steps increases.}
\label{fig:degroot_sweep_epoch6}
\end{figure}

A particularly informative comparison arises from DeGroot opinion dynamics. At Epoch~6, DeGroot attains a relatively high correlation ($\rho_6=0.846$), yet its terminal opinion diversity $\mathrm{Div}_6$ is extremely small (Table~\ref{tab:external_baselines}). DeGroot also obtains higher Acc@3 than our method ($0.361$ vs.\ $0.278$), so our advantage is not uniform across all metrics. The step sweep in Fig.~\ref{fig:degroot_sweep_epoch6} shows that diversity collapses sharply within the first few steps and remains near zero thereafter, while ranking performance continues to deteriorate. Our method is not directly compared on terminal diversity because Table~\ref{tab:external_baselines} does not report $\mathrm{Div}_6$ for the LLM-agent mechanism. The supported comparison is final ranking behavior under different mechanisms: DeGroot exhibits near-zero terminal diversity as its number of update steps increases, whereas our method achieves lower final $\mathrm{DPAE}$ under a different, locally constrained interaction mechanism.

\begin{table}[t]
\centering
\caption{External baselines at Epoch~6 on the Social-Observed subset ($n=419$). Lower $\mathrm{DPAE}=1-\rho$ is better; higher $\rho$ and Acc@3 are better; Div$_6$ reports DeGroot terminal opinion diversity.}
\label{tab:external_baselines}
\scriptsize
\setlength{\tabcolsep}{4pt}
\begin{tabular}{@{}lcccc@{}}
\toprule
Method & DPAE$_6$ & $\rho_6$ & Acc@3$_6$ & Div$_6$ \\
\midrule
Random & $0.999\pm0.147$ & 0.001 & 0.028 & -- \\
Self-Only & $1.059\pm0.159$ & -0.059 & 0.111 & -- \\
Linear Regression & $0.838\pm0.173$ & 0.162 & 0.083 & -- \\
MLP Regressor & $0.839\pm0.121$ & 0.161 & 0.083 & -- \\
\midrule
SGC/GCN-like (1-hop) & $0.977\pm0.179$ & 0.023 & 0.111 & -- \\
GAT-like (1-hop) & $0.847\pm0.228$ & 0.153 & 0.139 & -- \\
DeGroot dynamics & $0.154\pm0.044$ & 0.846 & $\bm{0.361}$ & $5\times10^{-4}$ \\
\midrule
Ours & $\bm{0.124\pm0.009}$ & \textbf{0.876} & 0.278 & -- \\
\bottomrule
\end{tabular}
\end{table}

\FloatBarrier

\subsection{Exploratory Classroom Heterogeneity}

\begin{figure}[t]
\centering
\includegraphics[width=0.98\textwidth,trim=8 0 0 4,clip]{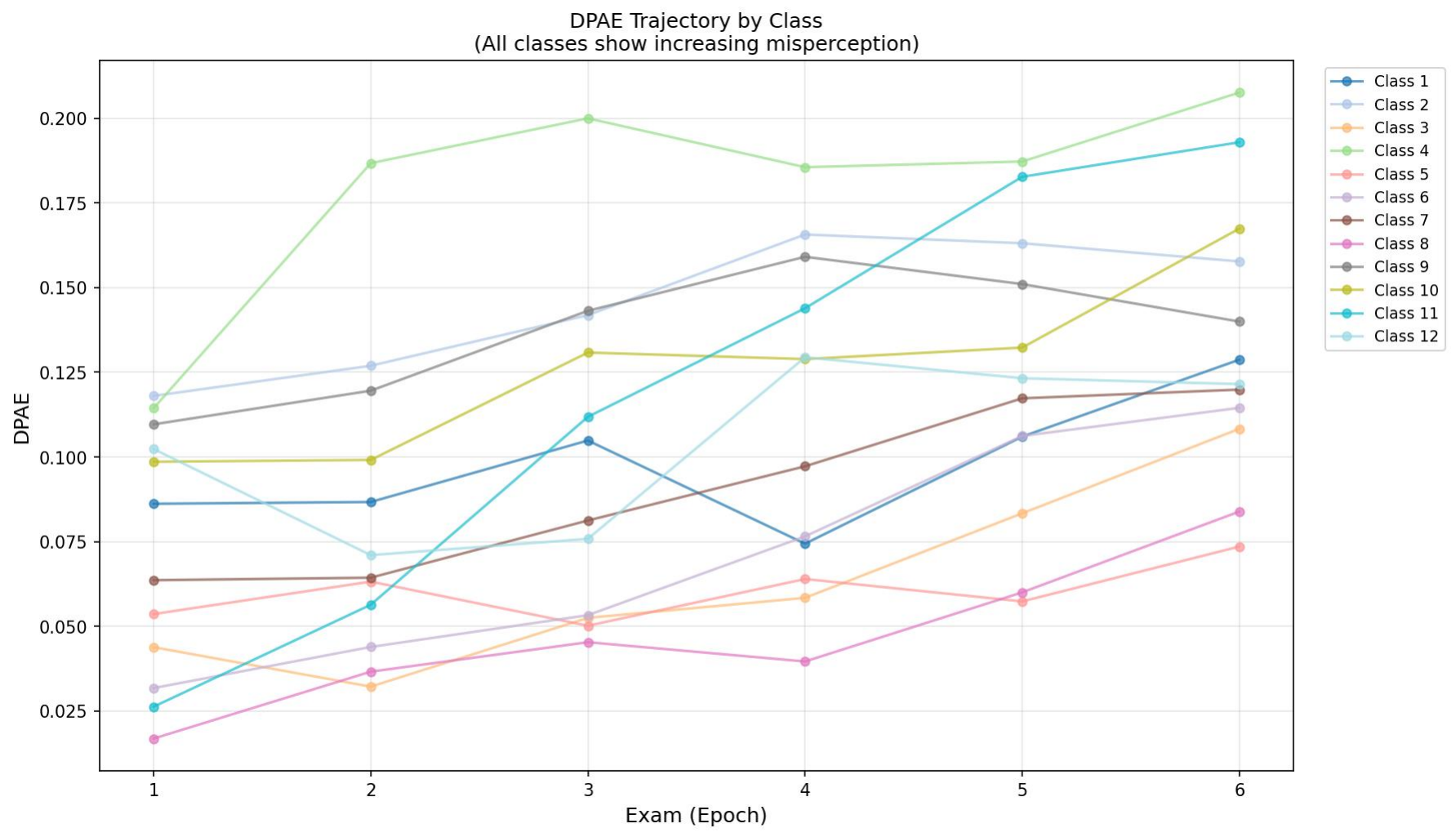}
\caption{Class-wise six-epoch $\mathrm{DPAE}$ trajectories. Lower is better; trajectories show that simulated misperception growth rates differ across classrooms.}
\label{fig:classwise_trajectories}
\end{figure}

At the classroom level, $\mathrm{DPAE}$ exhibits pronounced heterogeneity. Different classes end at markedly different misperception levels, and the growth rate of misperception varies across classes (Fig.~\ref{fig:classwise_trajectories}). This descriptive pattern shows that the simulation does not produce a single uniform trajectory across classrooms; identifying which classroom attributes account for the variation requires additional analysis.

\FloatBarrier

\section{Ethics and Data Governance}

The study uses sensitive educational records involving minors, including exam scores, peer-relation questionnaire responses, and social-anxiety-related variables. The secondary analysis of these pre-existing, de-identified records was reviewed and approved by an institutional ethics review committee and conducted under the applicable data-use authorization. School names, student names, class identifiers, and other direct identifiers were excluded from the analysis. Because classroom social networks may remain sensitive even after de-identification, we do not release raw student-level data.

\section{Conclusion}

This paper proposed a large-language-model-driven multi-agent probabilistic framework for data-constrained simulation of subjective classroom social perception. Departing from assumptions of global visibility or shared cognition, each agent is endowed with an individualized subjective social graph that constrains information acquisition, social interaction, and belief updating throughout the simulation. By integrating retrieval-augmented generation, trust gating, and probabilistic belief fusion, the framework provides a tractable way to study possible collective cognitive dynamics without relying on global information.

Experiments on middle-school classroom records show that, even when each individual receives an exogenous signal at every time step---the observed exam outcome---localized and noisy simulated social exchange can generate time-accumulating collective misperception. Ablation results suggest that subjective visibility and LLM-based trust gating are directionally associated with better long-horizon ranking behavior in the evaluated seeds. Compared with evaluated DeGroot configurations, the framework achieves lower final ranking error under a different locally constrained interaction mechanism; the DeGroot configurations exhibit near-zero terminal opinion diversity as update steps increase.

Methodologically, the contribution is not to introduce a stronger predictor, but to provide a modeling paradigm in which a language model acts as a \emph{structurally constrained cognitive module} embedded within a multi-agent system. The language model is not treated as an omniscient oracle; it is restricted to each agent's subjective view and locally retrievable evidence, making its outputs approximations of situated simulated judgment.

Several limitations remain. First, the experiments focus on classroom-scale systems, and scalability to larger networks has not yet been validated. Second, psychological variables are modeled through questionnaire-based mappings and parametric noise. Third, trust gating depends on LLM-generated evaluative scores, which may inherit self-preference or evaluator-specific biases despite structured prompts and constrained evidence. Finally, the present study is a descriptive mechanism simulation, not a causal validation of students' actual cognitive processes.

Future directions include extending the framework to other settings of social learning, incorporating richer behavioral traces, and exploring information interventions under constrained cognition.

\end{document}